\documentclass[11pt]{article}

\usepackage[top=1.0in, bottom=1.0in, left=1.0in, right=1.0in]{geometry}
\usepackage{amsmath}
\usepackage{amssymb}
\usepackage{amsthm}
\usepackage{bm}
\usepackage{mathrsfs}
\usepackage{graphicx}
\usepackage{epsfig}
 \usepackage{subfig}
\usepackage{paralist}
 \usepackage{cite}
\usepackage{setspace}
\usepackage{cancel} 
\usepackage{appendix}
\usepackage{color}
\usepackage{wrapfig}
\usepackage{xspace}
\usepackage[colorlinks, citecolor=black, linkcolor=black, linktocpage=true]{hyperref}
\usepackage{bookmark}
\usepackage{multirow} 
\usepackage{chngpage}
\usepackage{booktabs}
\usepackage{threeparttable}
\usepackage{epstopdf}

\newcommand{\ignore}[1]{}

\newcommand{\bma}[1]{\left[\begin{array}{#1}}
\newcommand{\ema}{\end{array}\right]}

\DeclareMathAlphabet{\mbf}{OT1}{ptm}{b}{n}
\newcommand{\mbs}[1]{{\boldsymbol{#1}}}

\newcommand{\mbshat}[1]{{\hat{\boldsymbol{#1}}}}

\newcommand{\mbfhat}[1]{{\hat{\mbf{#1}}}}

\newcommand{\mbftilde}[1]{{\tilde{\mbf{#1}}}}

\newcommand{\utimes}{{\raisebox{-0.6ex}{ \kern-1.0ex\raisebox{0.6ex}{\small$\mathsf{v}$}}} } %

\newcommand{\trans}{{\ensuremath{\mathsf{T}}}} 
\newcommand{\ura}[1]{{\underrightarrow{{#1}}}}

\def\fdotb{{\raisebox{-0.6ex}{ \kern0.2ex\raisebox{0.8ex}{\tiny $\hspace*{-1ex}\circ$}}}}

\def\fddotb{{\raisebox{-0.6ex}{ \kern0.2ex\raisebox{0.8ex}{\tiny $\hspace*{-1ex}\circ\circ$}}}}

\newcommand{\onehalf}{\mbox{$\textstyle{\frac{1}{2}}$}}

\newcommand{\beq}{\begin{equation}}
\newcommand{\eeq}{\end{equation}}
\newcommand{\bdis}{\begin{displaymath}}
\newcommand{\edis}{\end{displaymath}}
\newcommand{\beqarray}{\begin{eqnarray}}
\newcommand{\eeqarray}{\end{eqnarray}}
\newcommand{\beqarraynn}{\begin{eqnarray*}}
\newcommand{\eeqarraynn}{\end{eqnarray*}}

\makeatletter
\renewcommand{\p@enumii}{\theenumi.}
\makeatother

\theoremstyle{definition} 

\title{\textbf{Differential Geometric SLAM}}

\author{David Evan Zlotnik (Ph.D. Candidate) and James Richard Forbes (Assistant Professor) \\ Department of Aerospace Engineering, University of Michigan, Ann Arbor, MI, USA}
\date{\today}

\begin{document}

\maketitle

\vspace{-40pt}
\section{Introduction}

Simultaneous localization and mapping (SLAM) is a fundamental problem in robotics. It is the process of creating a map of the environment while, at the same time, estimating the position and attitude of the robot relative to the map. SLAM enables autonomous path planning and control.

The extended Kalman Filter (EKF), Particle Filters, and Expectation Minimization are among the most popular SLAM methods \cite{book_thrun_2005,paper_aulinas_2008}. The EKF and the extended Information Filter (EIF) involve propagating a covariance matrix, or its inverse as in the EIF, along with the states of the robot and map. This leads to computational inefficiency for large scale problems where the number of landmarks may be very large. Moreover, the process and measurement noise are assumed to be Gaussian within the EKF formulation. This may not be a problem for robotic systems equipped with high quality sensors. However, low-cost inertial measurement units (IMUs) are plagued by high levels of non-Guassian noise and biases \cite{paper_mahony_2008}. Consequently, the EKF SLAM algorithm can be difficult to apply to robotic systems equipped with low-cost IMUs. In addition, inherent linearization in the EKF formulation can lead to computational difficulty as well. 

We present a differential geometric SLAM (DG-SLAM) algorithm that evolves directly on the special Euclidean group, $SE(3)$. The proposed filter employs methods from differential geometry to propagate the state and map estimates. Differential geometric methods have been employed previously in the literature to the problems of state estimation on $SO(3)$ and $SE(3)$ \cite{paper_mahony_2008,paper_baldwin}. Unlike EKF SLAM, the proposed filter is provably asymptotically stable. That is to say that, in the absence of bias and noise, the estimated states are guaranteed to converge to the true states. Another advantage of the algorithm is the absence of matrix inversions, which makes the algorithm suitable for large-scale implementation. By approaching the SLAM problem in a geometric framework we hope to avoid the pitfalls of traditional SLAM techniques. In particular, we hope that the proposed algorithm is robust to non-Gaussian noise associated with low-cost sensors.

\section{Applications}

\begin{figure}[h!]
	\centering
	\includegraphics[width=0.3\linewidth]{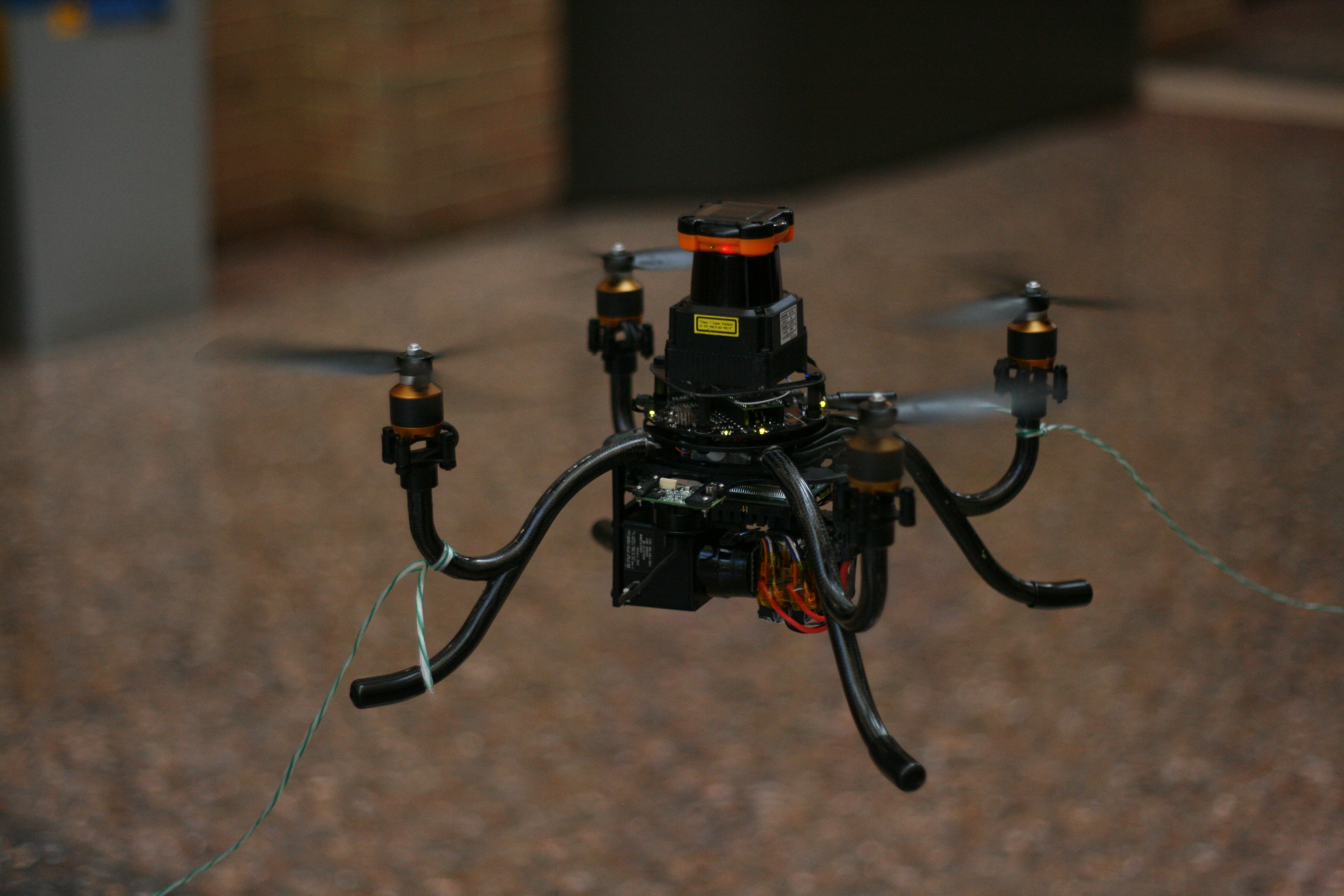}
	\includegraphics[width=0.3\linewidth]{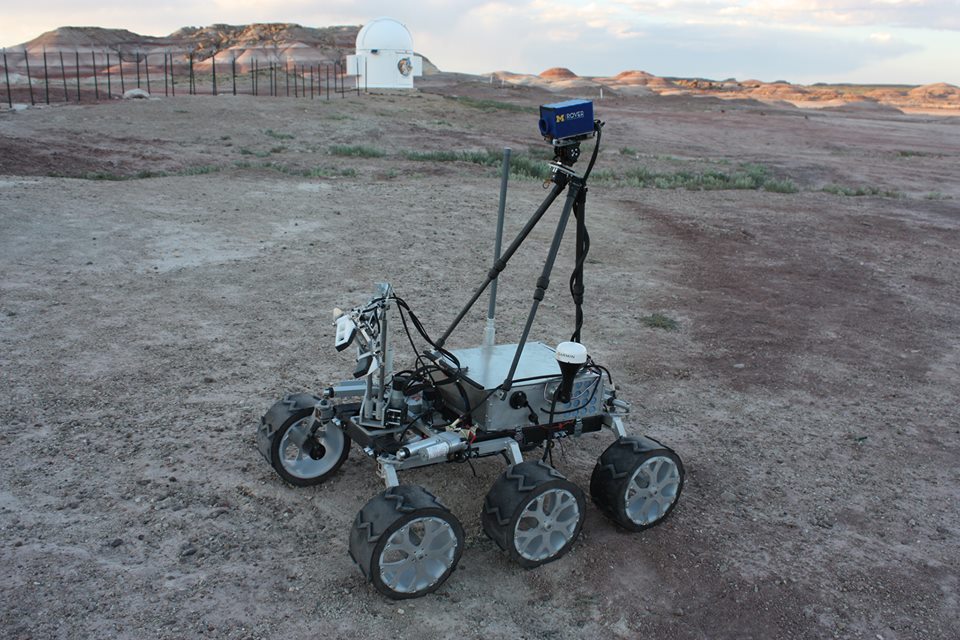} 
	\includegraphics[width=0.61\linewidth]{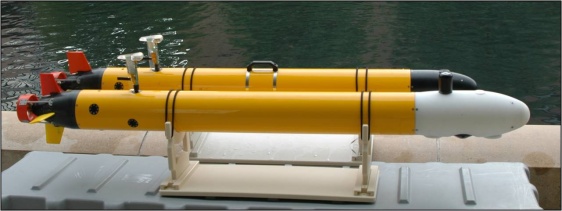} 
	\caption{UAVs, terrestrial robots, submersibles. Credit: MAAV (\url{http://www.maavumich.org}), MRover (\url{http://www.umrover.org}), PeRL (\url{http://robots.engin.umich.edu}).}
	\label{fig:vehicles}
\end{figure}

As shown in Figure \ref{fig:vehicles} the proposed DG-SLAM algorithm can be used to build a map and simultaneously localize a vehicle moving in three dimensional space in a variety of environments. 

\section{Notation}

\begin{figure}[ht!]
	\centering
	\vspace{-10mm}
	\includegraphics[width=0.45\textwidth]{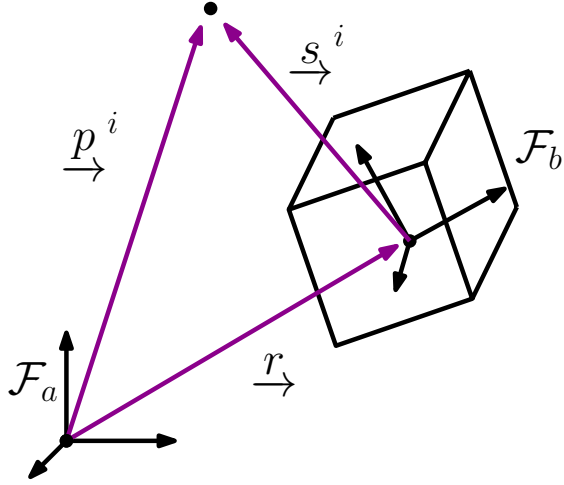}
	\caption{Vehicle, landmarks, and reference frames.}
	\label{fig:set_up}
\end{figure}

We consider the simultaneous mapping and localization problem in three dimensions. First let $\mathcal{F}_a$ and $\mathcal{F}_b$ denote the body-fixed frame of the robot and the datum frame, respectively, as shown in Figure~\ref{fig:set_up}. The goal is to estimate the pose of the robot and the landmark positions, relative to the datum frame. Let $\ura{r}$ denote the position of the center of mass of the vehicle relative to the origin of the datum frame. Also, let $\ura{p}^i$, $i \in \{ 1,\ldots,\ell\}$ denote the position of landmark $i$ relative to the origin of the datum frame. Vectors $\ura{r}$ and $\ura{p}^i$ may be resolved in any reference frame. For example, the representation of $\ura{r}$ and $\ura{p}^i$ in $\mathcal{F}_a$ is denoted $\mbf{r}_a$ and $\mbf{p}^i_a$, where the subscript denotes the frame in which the vector has been resolved. An additional vector, denoted $\ura{s}^i$ denotes the position of landmark $i$ relative to the center of mass of the vehicle.

The pose of the robot is described by $\mbf{X} \in SE(3)$, where 
\beq
	SE(3)  =  \left \{ \bma{c c} \mbf{C} & \mbf{r} \\ \mbf{0} & 1 \ema \in \mathbb{R}^{4\times4} \ | \ \mbf{C} \in \mathbb{R}^{3\times3}, \mbf{C}^\trans \mbf{C} = \mbf{1}, \text{det}(\mbf{C}) = +1, \mbf{r} \in \mathbb{R}^3 \right \}. \nonumber
\eeq
Let
\beq
	\mbf{X} = \bma{c c} \mbf{C}_{ba}^\trans & \mbf{r}_a \\ \mbf{0} & 1\ema \in SE(3),
\eeq
denote the true pose of the robot. The direction cosine matrix $\mbf{C}_{ba}$ maps the coordinates of a vector resolved in $\mathcal{F}_a$ to being resolved in $\mathcal{F}_b$. It is a representation of the attitude of the robot. The estimate of the pose of the robot is
\beq
	\mbfhat{X} = \bma{c c} \mbf{C}_{ea}^\trans & \mbfhat{r}_a \\ \mbf{0} & 1\ema \in SE(3),
\eeq
where $\mbf{C}_{ea}$ is the estimate of $\mbf{C}_{ba}$ and $\mbfhat{r}_a$ is the estimate of $\mbf{r}_a$.
Similarly, the true map is denoted $(\mbf{p}_a^1, \mbf{p}_a^2, \ldots, \mbf{p}_a^\ell)$, while the map estimate is denoted $(\mbfhat{p}_a^1, \mbfhat{p}_a^2, \ldots, \mbfhat{p}_a^\ell)$.

\section{Differential Geometric SLAM Formulation}

The following measurements are assumed available: (1) angular velocity, $\mbs{\omega}_b^y$, (2) velocity, $\mbf{v}_b^y$, and (3) the position of each landmark relative to the robot, $\mbf{s}_b^{i,y}$. All measurements are taken in the body frame of the vehicle $\mathcal{F}_b$. For the purpose of the DG-SLAM formulation and the stability proof, noise and bias associated with measurements is assumed zero. 

The proposed filter takes the following form:

\vspace{-2mm}
\begin{equation*}
\begin{aligned}[c]
	\dot{\mbfhat{X}}  &=  \mbfhat{X} \mbfhat{A} ,\ \  \mbfhat{A} = \bma{c c} \mbshat{\omega}^\times & \mbfhat{v} \\ \mbf{0} & 0 \ema, \\
	\dot{\mbfhat{p}}_i  &=  \mbf{C}_{ea}^\trans \mbs{\alpha}_i, 
	\end{aligned}
	\end{equation*}
	where
	\begin{equation*}
	\begin{aligned}[c]
	\mbshat{\omega}  &=  \mbs{\omega}_b^y - k_1 \mbf{e},\\
	\mbf{e} &= \onehalf ( \mbf{C}_{ba}\mbf{C}_{ea}^\trans - \mbf{C}_{ea} \mbf{C}_{ba}^\trans)^\utimes, \\
	\mbfhat{v}  &=  \mbf{v}_b^y + (\mbshat{\omega} - \mbs{\omega}_b^y)^\times \mbf{C}_{ea} \mbfhat{r}_a + k_2  \sum_{i=1}^\ell \mbftilde{s}^i - k_3 ( \mbf{C}_{ea} \mbfhat{r}_a + \mbf{s}_b^{1,y}),\\
	\mbs{\alpha}_i & =  (\mbshat{\omega} - \mbs{\omega}_b^y)^\times \mbf{C}_{ea} \mbfhat{p}_a^i - k_2  \mbftilde{s}^i,
\end{aligned}
\end{equation*}
and $\mbftilde{s}^i =  \mbf{C}_{ea} (\mbfhat{p}_a^i - \mbfhat{r}_a) - \mbf{s}_b^{i,y}$ is the innovation, which incorporates the error in the map and the pose.

\section{Stability Results}

By employing the Lyapunov function (candidate),
\beq
	V(\mbftilde{X},\mbftilde{p}_a^1,\ldots,\mbftilde{p}_a^\ell) = \onehalf || \mbf{1} - \mbftilde{X} ||_{\mathsf{F}}^2 + \sum_{i=1}^\ell ||\mbftilde{p}_a^i ||_2^2, \nonumber
\eeq
it can be shown that $\mbftilde{X} \rightarrow \mbf{1}$ and $\mbftilde{p}_a^i \rightarrow \mbf{0}$ as $t \rightarrow \infty$, where $\mbftilde{X} = \mbfhat{X} \mbf{X}^{-1}$ denotes the error in the pose, and $\mbftilde{p}_a^i = \mbf{C}_{ea} \mbfhat{p}_a^i - \mbf{C}_{ba} \mbf{p}_a^i$ denotes the error in the map. In the proof, it is assumed that $\mbf{C}_{ba}$ is known, however this is not true in practise. A similar problem occurs in similar $SO(3)$ and $SE(3)$ filters \cite{paper_mahony_2008,paper_baldwin}. In the works of \cite{paper_mahony_2008,paper_baldwin} a geometric approximation of $\mbf{C}_{ba}$, $\mbf{C}_{ba}^y = \mbf{C}_{ba}^y(\mbf{s}_b^{i,y}, \mbf{s}_a^i)$, is constructed from measurements and used in place of $\mbf{C}_{ba}$. In the proposed algorithm, $\mbf{C}_{ba}^y$ is constructed from landmark measurements and the estimated landmark positions (i.e., $\mbf{C}_{ba}^y = \mbf{C}_{ba}^y(\mbf{s}_b^{i,y}, \mbfhat{p}_a^i - \mbfhat{r}_a)$). When using $\mbf{C}_{ba}^y(\mbf{s}_b^{i,y}, \mbfhat{p}^i_a - \mbfhat{r}_a)$, the map estimate does not converge to the true map resolved in $\mathcal{F}_a$. However, the estimated positions of the landmarks relative to the estimated pose of the robot converges to their true counterparts. This is to be expected, since the global SLAM problem is unobservable unless at least two landmarks positions are known \cite{book_wang}.

\section{Simulation Results}

\begin{figure}[ht!]
	\centering
	\subfloat[$ t = 0$ ($s$)]{\includegraphics[width=0.4\textwidth]{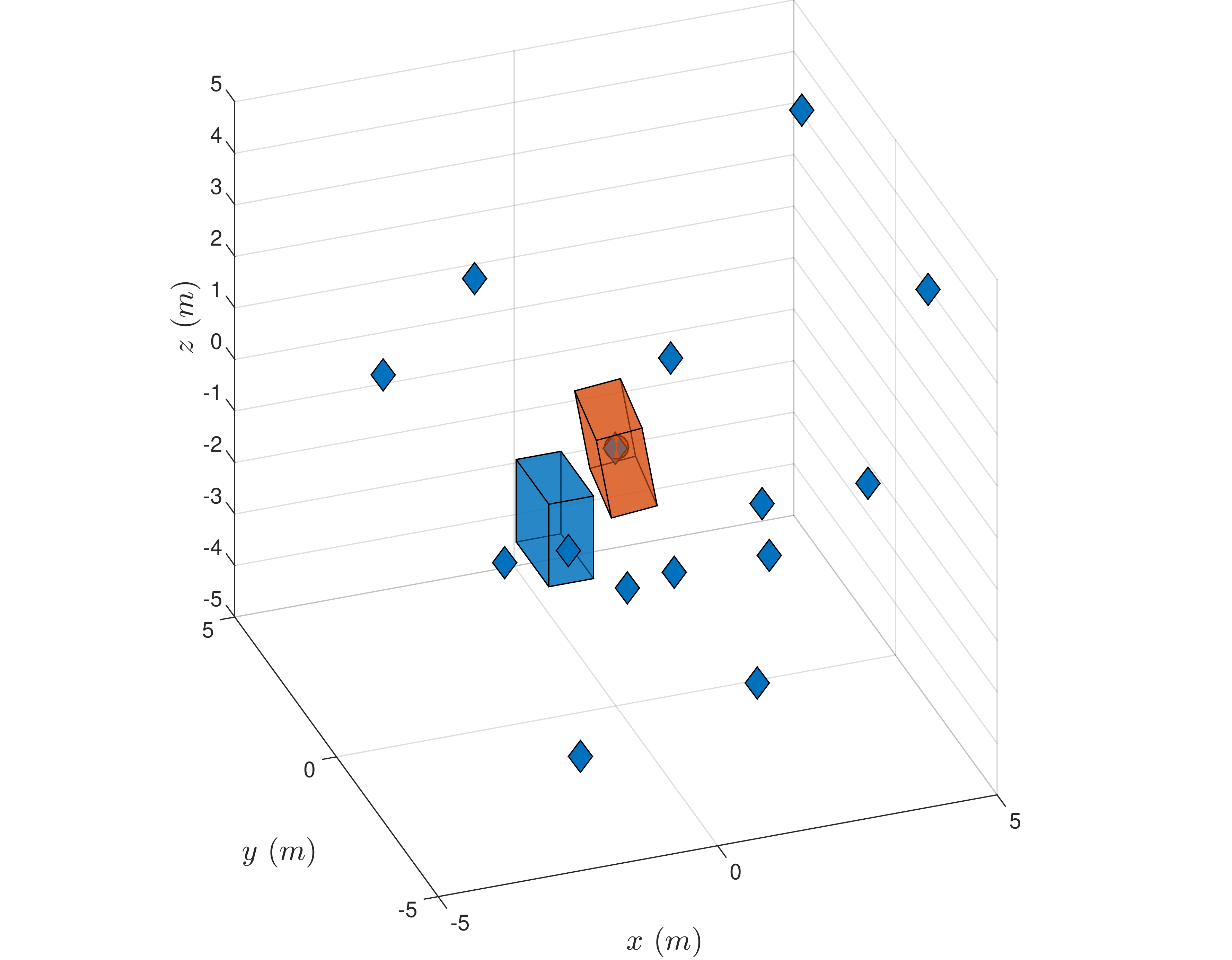}%
	\label{fig:t0}}
	\subfloat[$ t = 5$ ($s$)]{\includegraphics[width=0.4\textwidth]{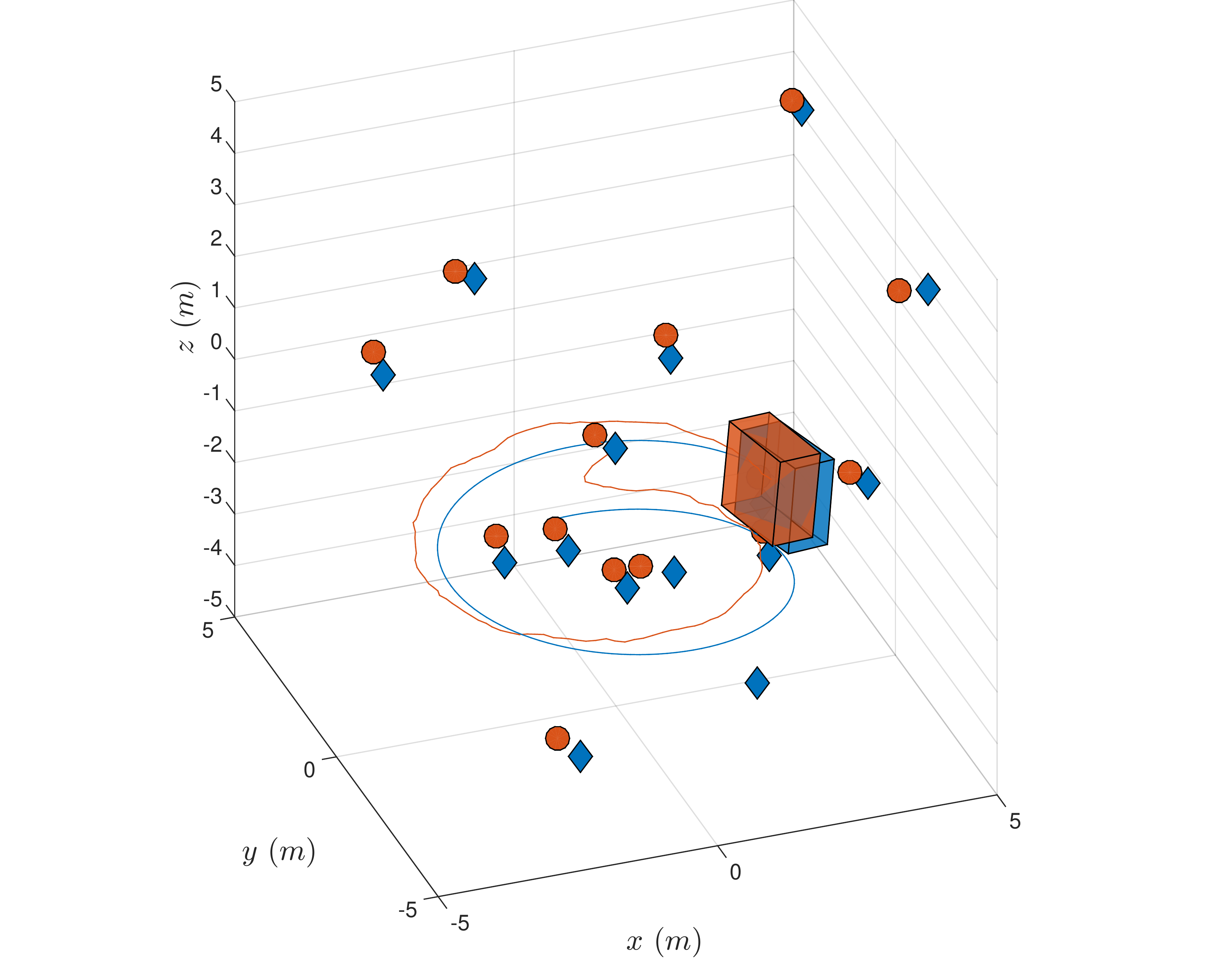}%
	\label{fig:t5}} \\
	\subfloat[$ t = 10$ ($s$)]{\includegraphics[width=0.4\textwidth]{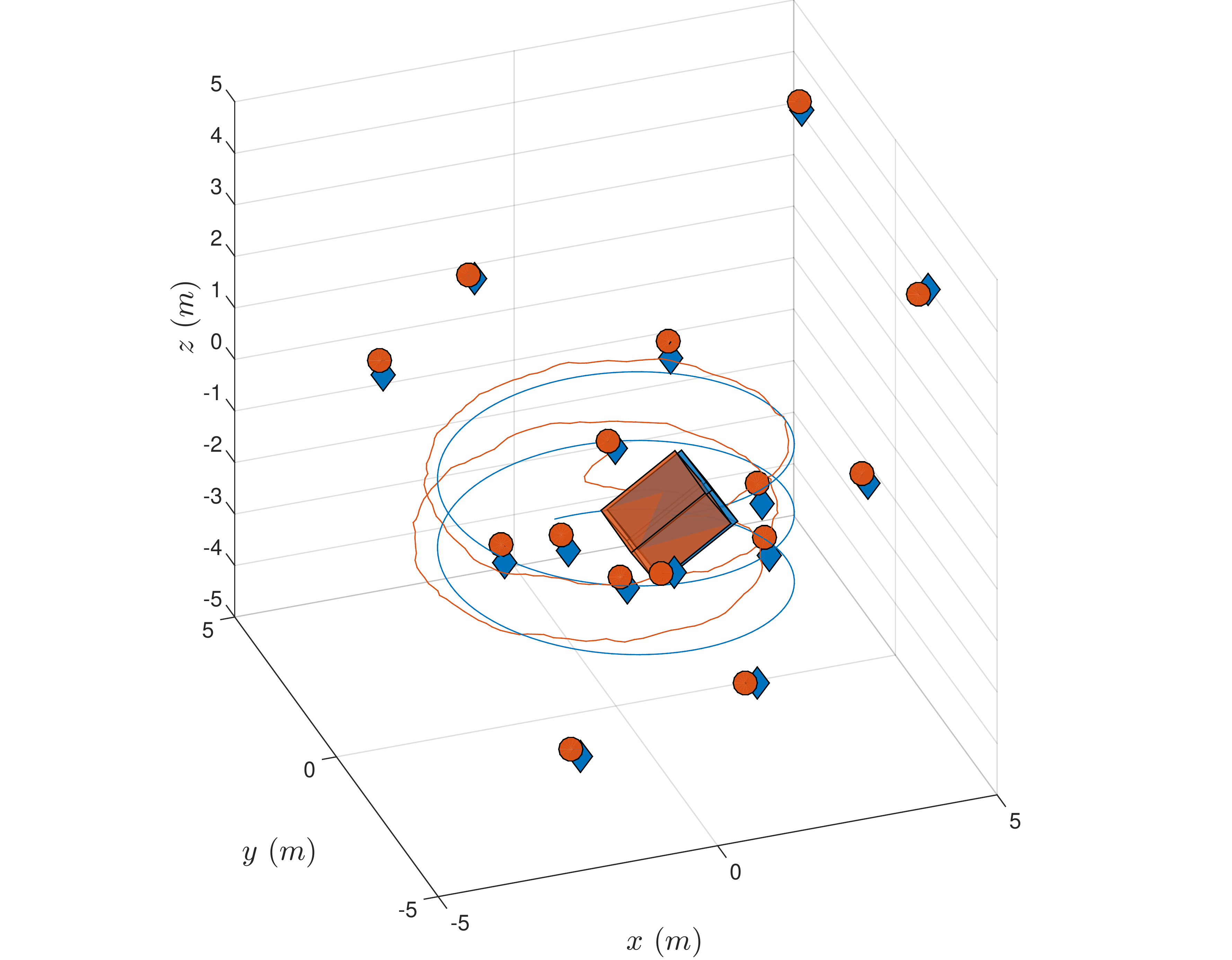}%
	\label{fig:t10}} 
	\subfloat[$ t = 20$ ($s$)]{\includegraphics[width=0.4\textwidth]{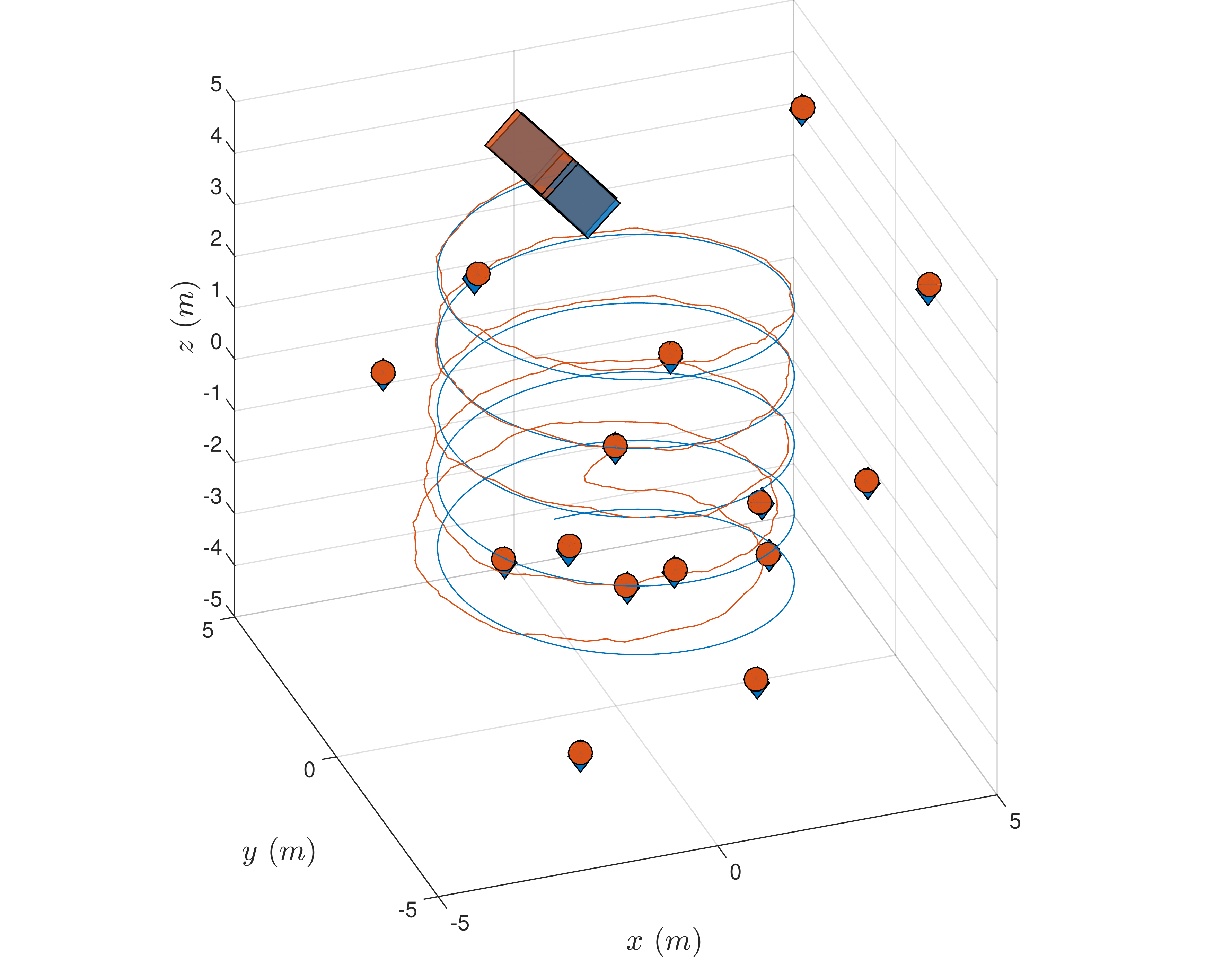}%
	\label{fig:t20}}
	\caption{ Visualization of SLAM algorithm. The blue diamonds denote the true positions of the landmarks. The orange markers denote the estimated landmark positions. The blue and orange boxes denote the true and estimated pose of the robot, respectively. During an initial period of 5 seconds the pose and map estimates rapidly converge towards the true pose and map. After 20 seconds the errors in the pose and map estimates have approached zero.}
	\label{fig:anim}
\end{figure}

\begin{figure}[ht!]
	\centering
	\vspace{-30mm}
	\includegraphics[width=0.45\linewidth]{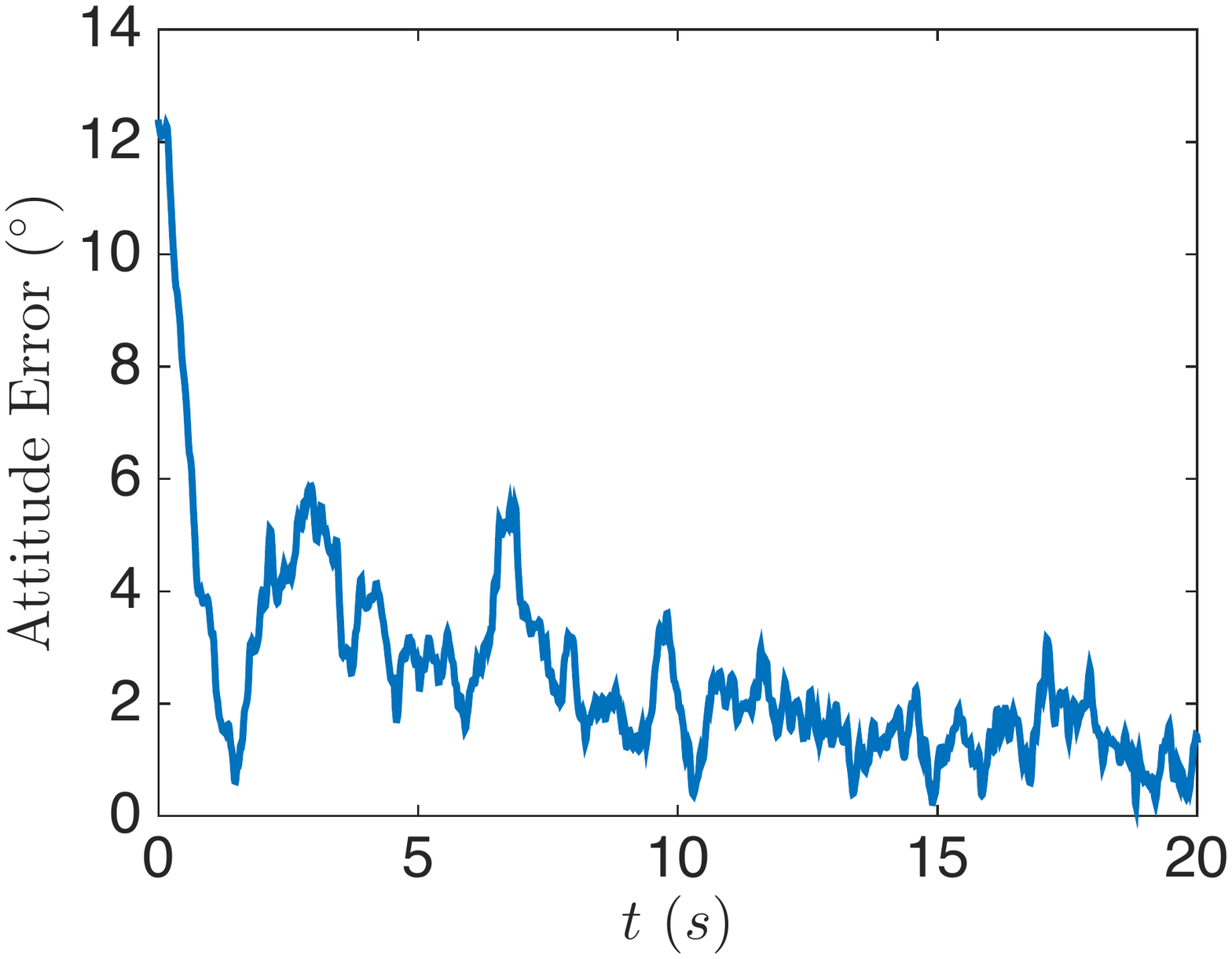} \qquad
	\includegraphics[width=0.45\linewidth]{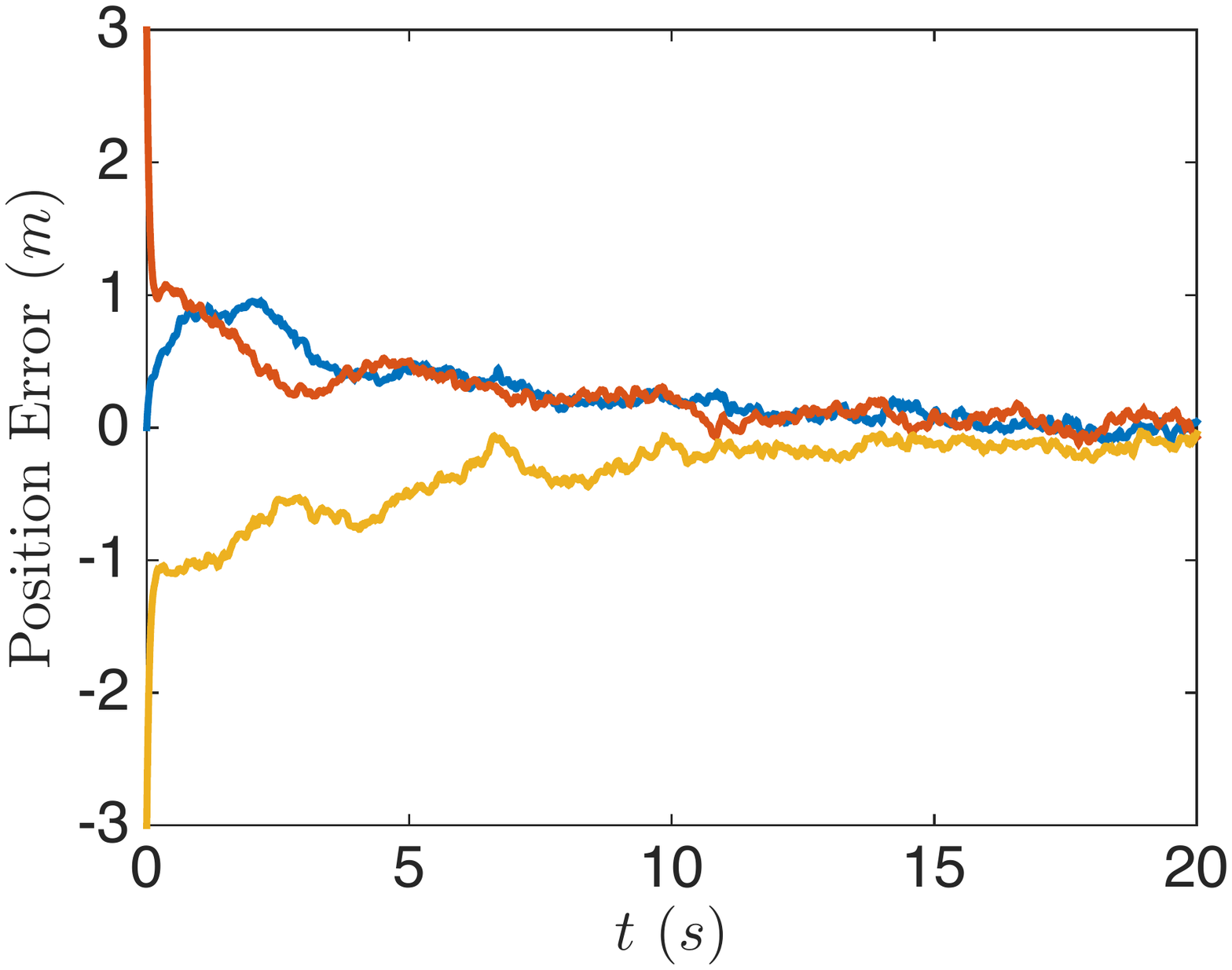} \\
	\vspace{-10mm}
	\includegraphics[width=0.5\linewidth]{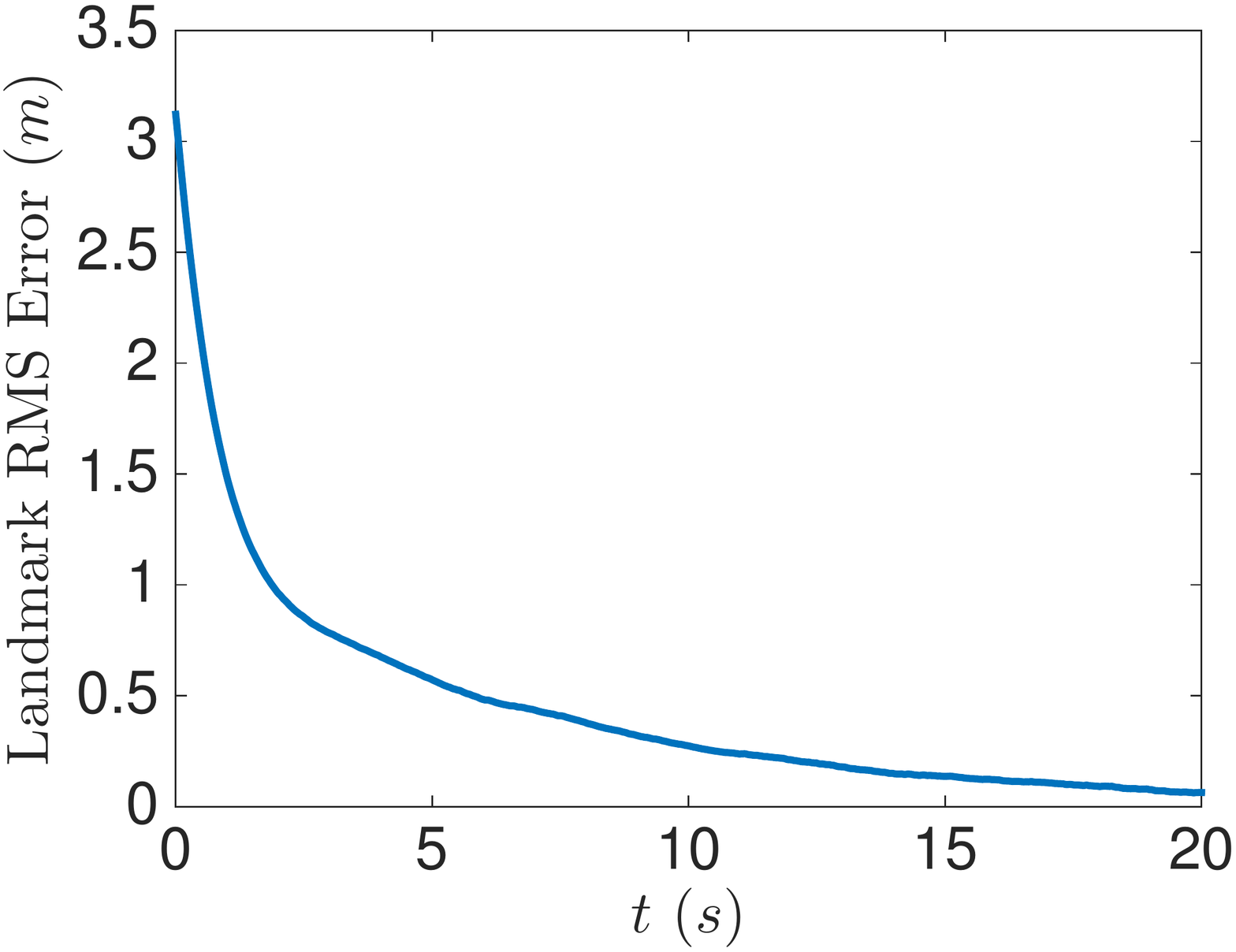}
	\vspace{-15mm}
	\caption{Simulation results. The error in the pose of the robot as well as the map approaches zero.}
	\label{fig:results}
\end{figure}

The DG-SLAM algorithm is employed in simulation. Although the stability proof assumes no noise, to assess robustness
of the proposed DG-SLAM algorithm, noise has been added to the angular
velocity, velocity, and vector measurements in the simulation. The results of the simulation can be found in Figures \ref{fig:anim} and \ref{fig:results}.

\section{Conclusions}

The proposed SLAM algorithm is based on differential geometric principles and is similar to $SO(3)$ and $SE(3)$ estimators found in the literature. The algorithm requires no matrix inversions making it suitable for large scale implementation. 
Moreover, the filter is guaranteed asymptotically stable, assuming no measurement noise, no bias, and $\mbf{C}_{ba}$ is known.

\bibliographystyle{aiaa}
\bibliography{DGSLAM}

\end{document}